\documentclass{article}

\usepackage[dblblindworkshop, final]{neurips_2025}
\workshoptitle{Scaling Environments for Agents (SEA)}

\usepackage[utf8]{inputenc} % allow utf-8 input
\usepackage[T1]{fontenc}    % use 8-bit T1 fonts
\usepackage{hyperref}       % hyperlinks
\usepackage{url}            % simple URL typesetting
\usepackage{booktabs}       % professional-quality tables
\usepackage{amsfonts}       % blackboard math symbols
\usepackage{nicefrac}       % compact symbols for 1/2, etc.
\usepackage{microtype}      % microtypography
\usepackage{xcolor}         % colors
\usepackage{graphicx}%
\usepackage{multirow}%
\usepackage{amsmath,amssymb,amsfonts}%
\usepackage{amsthm}%
\usepackage{bbm}
\usepackage[title]{appendix}%
\usepackage{xcolor}%
\usepackage{textcomp}%
\usepackage{manyfoot}%
\usepackage{booktabs}%
\usepackage{algorithm}%
\usepackage{algorithmicx}%
\usepackage{algpseudocode}%
\usepackage{listings}%
\usepackage{subcaption}
\usepackage{enumitem}
\usepackage{tcolorbox}

\newtheorem{prop}{Proposition}[section]

\theoremstyle{definition}
\newtheorem{definition}{Definition}[section]

\title{On the Importance of Task Complexity in Evaluating LLM-Based Multi-Agent Systems}

\author{
  \textbf{Bohan Tang\textsuperscript{1}\thanks{Equal contribution.},} 
  \textbf{Huidong Liang\textsuperscript{1}$^{*}$,} 
  \textbf{Keyue Jiang\textsuperscript{2}$^{*}$,} 
  \textbf{Xiaowen Dong\textsuperscript{1}}
\\
  \textsuperscript{1}University of Oxford, \textsuperscript{2}University College London \\
\texttt{bohan.tang@eng.ox.ac.uk}\\
}

\begin{document}

\maketitle

\begin{abstract}
    Large language model multi-agent systems (LLM-MAS) offer a promising paradigm for harnessing collective intelligence to achieve more advanced forms of AI behaviour. While recent studies suggest that LLM-MAS can outperform LLM single-agent systems (LLM-SAS) on certain tasks, the lack of systematic experimental designs limits the strength and generality of these conclusions. We argue that a principled understanding of task complexity, such as the degree of sequential reasoning required and the breadth of capabilities involved, is essential for assessing the effectiveness of LLM-MAS in task solving. To this end, we propose a theoretical framework characterising tasks along two dimensions: depth, representing reasoning length, and width, representing capability diversity. We theoretically examine a representative class of LLM-MAS, namely the multi-agent debate system, and empirically evaluate its performance in both discriminative and generative tasks with varying depth and width. Theoretical and empirical results show that the benefit of LLM-MAS over LLM-SAS increases with both task depth and width, and the effect is more pronounced with respect to depth. This clarifies when LLM-MAS are beneficial and provides a principled foundation for designing future LLM-MAS methods and benchmarks.
\end{abstract}

\section{Introduction}
\label{sec:intro}
Recent advances in AI technologies, exemplified by autonomous agents powered by large language models (LLMs), offer an exciting opportunity to explore the notion of a future society where machines assist and interact with humans in a wide range of tasks~\cite{li2023camel,rawles2024androidworld,tang2024synthesizing,luo2025vimo,tang2025beyond}. %The likely increasing presence of autonomous agents in both offline and online settings has the potential to improve efficiency and decision-making in human society. 
Rather than operating in isolation, multiple specialised agents can be deployed together, communicating and collaborating to address complex challenges. Consequently, LLM multi-agent systems (LLM-MAS) represent a promising paradigm for leveraging collective intelligence to achieve advanced forms of AI behaviours \cite{tran2025multi, li2024survey}. Understanding the effectiveness of such systems in task-solving is key to their deployment.

Current effort for understanding LLM-MAS in task solving predominantly focuses on empirically assessing the capability of the system in solving a wide range of tasks, including both discriminative (e.g., problem solving) and generative tasks (e.g., creative writing)~\cite{du2023improving,yu2025multi,huot2025agents}. Given a particular task, there is an increasingly large literature that investigates how better task-solving performance can be achieved via the optimisation of prompt engineering, as well as communication networks among AI agents~\cite{zhou2025multi,zhangg,zhuge2024gptswarm,yemas,zhang2025gnns}. While recent empirical results have shown that LLM-MAS can outperform LLM single-agent systems (LLM-SAS) on certain tasks, these findings are based solely on downstream performance.%, which means that any conclusions drawn are likely to be influenced by various aspects of the experimental setup. 
Therefore, we still lack a principled understanding of when and why an LLM-MAS system can be effective, i.e., it can outperform an LLM-SAS in task-solving. %\footnote{Notably, in this work, we focus on a representative class of LLM multi-agent systems, namely the LLM multi-agent debate system. Unless otherwise specified, all references to LLM multi-agent systems hereafter refer to this debate-based setting.}

\begin{figure}[t]
    \centering
    \includegraphics[width=\textwidth]{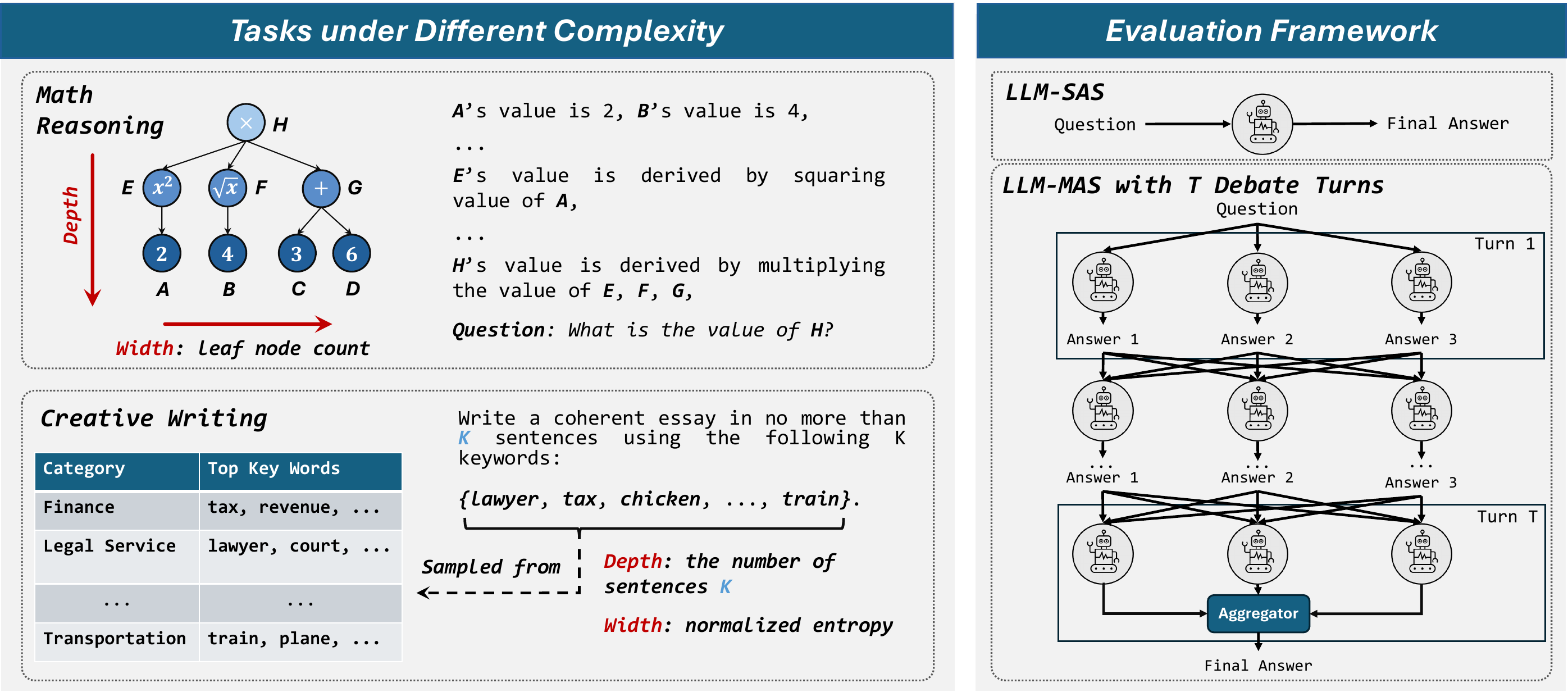} 
    \caption{\textbf{Left}: math reasoning and creative writing tasks with controllable complexity in terms of width and depth. \textbf{Right}: Exemplar LLM-SAS and LLM-MAS framework. For simplicity, the input question to each agent is omitted from the presentation starting with the second debate turn.}
    \label{fig:framework}
\end{figure}

In this work, we argue that a deeper understanding of the task complexity is the key to assessing the effectiveness of LLM-MAS in task-solving. Specifically, we focus on a representative class of LLM-MAS, namely a multi-agent debate system, in which multiple LLM agents collaboratively address a single problem, exchanging both their reasoning pathways and solution proposals throughout the process~\cite{du2023improving,zhangg,zhuge2024gptswarm,yemas}. Drawn from the literature on complex systems and collective intelligence~\cite{padhee2024echo,straub2023cost}, we introduce a two-dimensional measure of task complexity: \textbf{depth} and \textbf{width}. The depth of a task corresponds to the length of the reasoning chain an agent must follow, namely, the number of sequential inference or problem-solving steps required. The width of a task corresponds to the breadth of knowledge or skills required at each step, i.e., the range of capabilities that the agent needs to draw on per step. Figure~\ref{fig:framework} illustrates how LLM-SAS and LLM-MAS address tasks with varying levels of width and depth. Based on this two-dimensional measure, we propose a novel \textit{theoretical framework} to analyse how both task width and depth influence the performance of an LLM-MAS. We further demonstrate that, relative to the increase in task width, an increase in task depth tends to yield greater performance gains for LLM-MAS over LLM-SAS.

We empirically validate our theoretical findings in two representative tasks: math reasoning (a discriminative task) and creative writing (a generative task). First, in the math reasoning task, we employ the framework provided by the DyVal benchmark~\cite{zhu2024dyval} to dynamically adjust the task complexity, where the depth and width are naturally defined as those of the tree-based directed acyclic graph (DAG) used to generate the problem. Second, in the creative writing task, we propose a novel benchmark, named Depth-Width Writing (DW$^2$) benchmark, where the depth and width are the number of coherent sentences that are required to be written and the Shannon entropy of the domains of keywords to be used, respectively. Across both tasks, we observe two consistent patterns: 1) the benefit of LLM-MAS over LLM-SAS increases with task complexity; and 2) gains in task depth appear more significant compared to those in task width. These results affirm our theoretical findings and our main argument: a deeper understanding of the task complexity is the key to assessing the effectiveness of LLM-MAS in task-solving. Based on this insight, we discuss the open challenges and opportunities for designing future LLM-MAS. Our contributions are summarised as follows:

$\bullet$ We propose a novel theoretical framework to analyse the effectiveness of LLM-MAS. The framework is grounded in a two-dimensional measure of task complexity based on \textbf{depth} and \textbf{width}, which capture the degree of sequential reasoning and breadth of capabilities required for a task.

$\bullet$ We propose a simple framework to theoretically show that the performance gain of an LLM-MAS over an LLM-SAS grows with task complexity. Crucially, our analysis reveals that this advantage is more sensitive to increases in task depth than to that in task width.

$\bullet$ We provide extensive empirical validation for our theoretical findings in two representative tasks: math reasoning (a discriminative task) and creative writing (a generative task). For the latter, we propose a novel Depth-Width Writing (DW$^2$) benchmark, where depth is the number of coherent sentences required and width is the Shannon entropy of keyword domains. The empirical results affirm our theoretical findings and offer insights for designing more efficient LLM-MAS.

%We test the performance of LLM multi-agent task solving based on the multi-agent debate framework proposed by \cite{du2023improving}. In this framework, at each round, LLM agents independently attempt a task, and show their solution and reasoning process to their peers. This implicit debating process will prompt the agents to revise their solution at the next round. In both math reasoning and creative writing, we deploy an additional LLM agent as a summariser to generate the final answer for the multi-agent system. We then compare the performance of this multi-agent task solving system with that of a single-agent system, and importantly investigate the relation between performance gap and task complexity in terms of both task depth and width. 

% A series of exploratory experiments demonstrate that, for both types of tasks, those that can be better solved by a multi-agent system have complexity measures that are statistically significantly higher than those well-solved by a single-agent system. This validates task complexity, when it can be defined naturally or approximately, as an important angle through which we assess the effectiveness and necessity of multi-agent task solving.
\section{Analysis Framework}
\label{sec:analysis}
In this section, we develop a theoretical framework for analysing the effectiveness of LLM-MAS. We first set up our problem formulation in Subsection~\ref{subsec:pre}. Subsequently, in Subsection \ref{subsec:task_com}, we show how both task width and task depth affect the performance gain of LLM-MAS relative to LLM-SAS. Our analysis reveals that although increasing either dimension improves such performance gain of LLM-MAS, task depth plays a more decisive role than task width.

\subsection{Problem Formulation}
\label{subsec:pre}
\textbf{LLM multi-agent systems}. We define LLM multi-agent systems (LLM-MAS) as computational systems comprised of multiple autonomous agents, each empowered by LLMs, that perceive, reason, and collaborate to accomplish complex tasks at scale \cite{guo2024large, tran2025multi, li2024survey, ye2025maslab}. Unlike the LLM single-agent systems (LLM-SAS), where only one LLM operates in isolation, LLM-MAS utilises natural language as a communication medium, enabling agents to coordinate, share context, and jointly refine their outputs toward a common goal. In this work, we focus specifically on the multi-agent debate system, a particularly prominent instantiation of LLM-MAS. In this framework, multiple LLM agents engage in proposing, critiquing, and refining arguments~\cite{du2023improving,yu2025multi,huot2025agents}.

\textbf{Problem statement.} While LLM-MAS offer the potential to improve problem-solving through collaboration, they also introduce additional computational overhead due to multi-agent coordination and repeated communication rounds~\cite{gao2025single}. Theron, we formalise effectiveness as the relative performance gain of LLM-MAS over its LLM-SAS counterparts on a given task. Our research problem can be formulated as investigating how intrinsic task properties influence the comparative effectiveness of LLM-MAS, to establish a principled understanding of when and why LLM-MAS shows advantages.

\subsection{Task Complexity for LLM-MAS Effectiveness}
\label{subsec:task_com}

In this subsection, we first introduce a two-dimensional measure for characterising task complexity in terms of \textbf{depth} and \textbf{width}. We then formalise the success rates of both LLM-SAS and LLM-MAS. Finally, we provide a theoretical analysis showing that: 1) increasing either task width or depth enhances the effectiveness of LLM-MAS relative to their single-agent counterparts; and 2) compared with width, increases in task depth generally yield greater performance gains for LLM-MAS.

\begin{figure}[t]
    \centering
    \includegraphics[width=\textwidth]{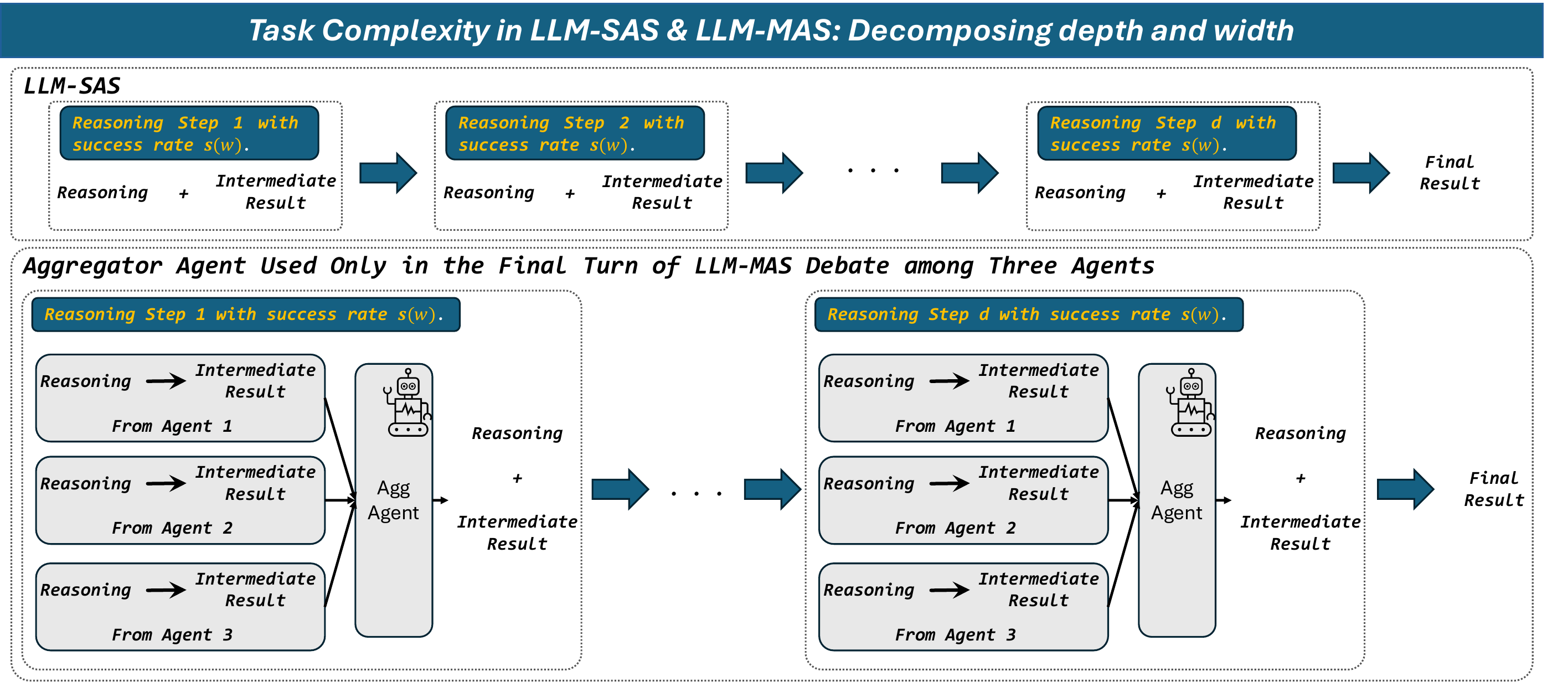} 
    \caption{Visualization of task complexity defined by depth and width. The pipeline represents one round of multi-agent debate. ``Agg'' stands for aggregator. }
    \label{fig:depth-vs-width}
\end{figure}

\textbf{Width and depth}. Building on prior works in complex systems and collective intelligence~\cite{padhee2024echo,straub2023cost}, we formalise task complexity using a two-dimensional measure defined by \textbf{depth} and \textbf{width}. The \emph{depth} of a task characterises the length of the reasoning chain an agent must traverse, that is, the number of sequential inference or problem-solving steps required. In contrast, the \emph{width} of a task captures the breadth of knowledge, skills, or alternatives that must be considered at each step, i.e., the range of capabilities an agent must draw upon. Figure~\ref{fig:depth-vs-width} provides a visualisation for task depth and width under LLM-SAS and LLM-MAS, which we will introduce as follows. 

\begin{definition}[The task defined by depth and width]
\label{def:dep_wid}
Consider a task requiring a reasoning chain of depth \(d \in \mathbb{N}\) (i.e., $d$ steps). At each step, the agent must complete \(w \in \mathbb{N}\) capability-specific micro-operations, which capture the width of the task\footnote{While the width can vary across reasoning steps, we assume the width to be fixed across steps for simplicity.}. Let the set of involved capabilities at a step be indexed by \(j = 1, \dots, w\), and denote by \(q_j \in (0,1)\) the success rate associated with capability \(j\). Since tasks relying on the same capability have highly correlated outcomes, we treat the success of each capability as a single event, independent across different capabilities. The per-step success rate is thus $s(w) \triangleq \prod_{j=1}^w q_j$. Theron, the task can be formalized as the sequence \(\mathcal{T} = \{s(w)\}_{t=1}^d\).
\end{definition}
% \begin{definition}[A task defined by depth and width]
% \label{def:dep_wid}
% Consider a task requiring a reasoning chain of depth \(d \in \mathbb{N}\) (i.e., $d$ steps). At each step, the agent must complete \(w \in \mathbb{N}\) capability-specific micro-operations, which capture the width of the task\footnote{While the width can vary across reasoning steps, we assume the width to be fixed across steps for simplicity.}. Let the set of involved capabilities at a step be indexed by \(j = 1, \dots, w\), and denote by \(q_j \in (0,1)\) the success rate associated with capability \(j\). Since tasks relying on the same capability have highly correlated outcomes, we treat the success of each capability as a single event, independent across different capabilities. The per-step success rate is thus $s(w) \triangleq \prod_{j=1}^w q_j$. %Hence, a single round of a multi-agent debate task can be characterised by a sequence of success rate \(\mathcal{T} = \{s(w)\}_{t=1}^d\), encapsulating both its depth and width. Once each agent has completed a reasoning step, their individual outputs are collected by an aggregation agent. This agent then produces a summary prompt to initiate the next step of reasoning.
% \end{definition}

To illustrate the intuition behind Definition~\ref{def:dep_wid}, consider the two tasks in Figure~\ref{fig:framework}. In the math reasoning task, depth is the number of steps along the reasoning chain, and width corresponds to applying different operators (e.g., addition, multiplication) within a step. Success on addition is independent of success on multiplication, so the per-step probability is the product of their accuracies. In the creative writing task, depth is the number of coherent sentences to be written, while width corresponds to correctly integrating each required keywords. Keywords from different domains are treated as independent, since correctly using an astronomy keyword does not make it easier or harder to integrate a culinary keyword. Based on this definition, we formalise the success rates of LLM-SAS and LLM-MAS. Our formulation focuses on the final round of discussion among the LLM agents. In this round, the answers produced by individual agents are summarised by one aggregator agent. Each agent conditions its response on both its own previous output and the prior responses of others, but the subsequent text generation is executed independently by the underlying LLM. Accordingly, we assume that the final-round outputs of different agents are independent. An illustration of such independent generation is provided in turn $T$ of LLM-MAS in Figure~\ref{fig:framework}. Moreover, for simplicity in subsequent derivations, we assume \(q_j = q\) for all \(j\), so that \(s(w) = q^w\).

\begin{definition}[The success rate of LLM-SAS]
\label{def:sr_llm_sas}
For a given task defined as \(\mathcal{T} = \{s(w)\}_{t=1}^d\). A single agent must correctly execute every step. Therefore, the success rate of an LLM-SAS across depth \(d\) with width \(w\) can be defined as $S_{\mathrm{single}}(d,w) \;\triangleq\; \bigl[s(w)\bigr]^{d}$.
\end{definition}

\begin{definition}[The success rate of LLM-MAS]
\label{def:sr_llm_mas}
For a given task defined as \(\mathcal{T} = \{s(w)\}_{t=1}^d\). Suppose we deploy \(N \ge 2\) agents and an aggregator agent that, upon observing the outputs, selects the correct answer with probability \(r \in (0,1]\) whenever at least one agent produced it. As a result, the success rate of an LLM-MAS across depth \(d\) with width \(w\) can be defined as $S_{\mathrm{multi}}(d, w, N, r) \;\triangleq\; r\,\bigl[1 - \bigl(1 - s(w)\bigr)^{N}\bigr]^{d}.
$
\end{definition}

% \textbf{Theoretical analysis.} Based on the Definition~\ref{def:sr_llm_sas} and Definition~\ref{def:sr_llm_mas}. We first formalise the LLM-MAS performance gain over LLM-SAS as follows.

\begin{definition}[LLM-MAS Performance Gain over LLM-SAS]
\label{def:pg}
For the scenarios with the LLM-MAS outperforming the LLM-SAS, we define the performance gain of the LLM-MAS over the LLM-SAS as the following relative-improvement function: 
$\Delta(d,w,N,r) \;\triangleq\; \frac{S_{\mathrm{multi}}-S_{\mathrm{single}}}{S_{\mathrm{single}}}.$
\end{definition}

\textbf{Theoretical analysis.} With Definition~\ref{def:pg}, we establish two key propositions to capture the respective roles of task width and depth in shaping the performance gain of LLM-MAS over LLM-SAS. The proofs for these propositions can be found in the Appendix.

\begin{prop}[Increase of LLM-MAS Performance Gain with Depth \(d\) and Width \(w\)]
\label{prop:grad_d_positive}
\setlength{\abovedisplayskip}{4pt}
\setlength{\belowdisplayskip}{4pt}
Let \(\mathcal{T} = \{s(w)\}_{t=1}^d\) denote a given task. According to Definition \ref{def:pg}, we have $\Delta(d,w,N,r) \;\triangleq\; \frac{S_{\mathrm{multi}}-S_{\mathrm{single}}}{S_{\mathrm{single}}}.$ 
Then, we have $\frac{\partial \Delta}{\partial d}\;>\;0$ and $\frac{\partial \Delta}{\partial w}\;>\;0$.
\end{prop}

Proposition~\ref{prop:grad_d_positive} demonstrates that increasing either dimension of task complexity, depth or width, amplifies the relative benefit of LLM-MAS over LLM-SAS. We attribute this to two complementary mechanisms: 1) increasing width lowers the per-step accuracy of any individual agent, thereby heightening the value of redundancy and error mitigation through multiple cooperative agents; and 2) increasing depth compounds this effect across a longer chain of reasoning steps, making single-agent error accumulation more detrimental, while LLM-MAS benefit from collaborative coverage to counteract this compounding failure risk. Consequently, both dimensions independently enhance the performance gain of LLM-MAS over LLM-SAS.

\begin{prop}[Unbounded Growth in Depth vs. Finite Saturation in Width]
\label{prop:limits_depth_width}
\setlength{\abovedisplayskip}{4pt}
\setlength{\belowdisplayskip}{4pt}
Let \(\mathcal{T} = \{s(w)\}_{t=1}^d\) denote a given task. According to Definition \ref{def:pg}, we have 
$\Delta(d,w,N,r) \;\triangleq\; \frac{S_{\mathrm{multi}}-S_{\mathrm{single}}}{S_{\mathrm{single}}}.$ Then, we have $\lim\limits_{w \to +\infty}\Delta(d,w,N,r)
\;=\;(rN)^{d}-1$ and $\lim\limits_{d\to+\infty}\Delta(d,w,N,r)
\;=\;+\infty$.
\end{prop}

Proposition~\ref{prop:limits_depth_width} highlights an asymmetry between the proposed two dimensions of task complexity. As the width grows, the benefit of having multiple agents eventually saturates: LLM-MAS cannot gain more than the \(N\)-fold redundancy adjusted by the aggregator’s reliability \(r\). By contrast, as depth grows, the advantage of LLM-MAS scales without bound. Intuitively, this arises because LLM-SAS suffer from compounding errors that degrade performance exponentially across long reasoning chains, whereas LLM-MAS can mitigate this compounding effect by distributing intermediate reasoning across agents and cross-validating partial results.

Together, Propositions~\ref{prop:grad_d_positive} and~\ref{prop:limits_depth_width} establish our central claim: both width and depth improve the relative performance of LLM-MAS, but depth exerts a greater influence. %While width-driven gains eventually plateau, depth-driven gains can potentially grow unboundedly with the number of reasoning steps required by a task. 

\section{Related Work}

\textbf{Analysis of LLM-MAS effectiveness.} 
The effectiveness of LLM-MAS has been empirically investigated in many recent work. \cite{yang2025revisiting} conceptualises LLM-MAS as a form of test-time computational scaling and finds that, while LLM-MAS yields limited benefits over strong LLM-SAS baselines in mathematical reasoning, its benefit grows when tasks become more difficult, model capabilities are limited, and agent diversity contributes minimally. \cite{gao2025single} demonstrates that as foundation models improve in reasoning and tool use, the performance gains from LLM-MAS diminish, motivating a hybrid agentic paradigm that selectively uses LLM-MAS and LLM-SAS for different user requests. In contrast, \cite{cemri2025multi} provides a systematic diagnosis of LLM-MAS performance limitations, introducing the Multi-Agent System Failure Taxonomy (MAST), identifying fourteen distinct failure modes. We complement and extend these empirical and diagnostic studies by introducing a theoretical framework linking LLM-MAS effectiveness directly to task complexity, characterised in two dimensions: depth (reasoning chain length) and width (the breadth of capabilities required per step). We further prove that performance gains from increasing width saturate, whereas those from increasing depth can grow unbounded, offering principled insight into when and why LLM-MAS outperform LLM-SAS.

\section{Experiments}
Following the theoretical findings introduced in Section~\ref{sec:analysis}, we empirically evaluate the behaviour of LLM-MAS and LLM-SAS under different task complexities.

\subsection{Experiment Setup}
We conduct extensive experiments across two complementary tasks: discriminative math reasoning and generative creative writing, as detailed below.

\textbf{Math reasoning dataset.} We consider the DyVal benchmark \cite{zhu2024dyval}, a synthetic math reasoning dataset in which problems are defined via a tree-based directed acyclic graph (DAG). As illustrated in Figure~\ref{fig:framework}, the root nodes denote the values of variables, where the other nodes represent mathematical operations $\{+,-,\times,\cdot^2,\sqrt{\cdot}\}$. The DAGs are later described in natural language with different task options, and we adopt the \texttt{linear equation} problems in our experiment. The task complexity naturally resides in the DAG structure, where \textbf{\textit{depth}} (i.e., number of reasoning steps) governs the sequential breakdown of sub-problems, and \textbf{\textit{width}} (i.e., number of children per node) controls the variety of operations required at each sub-problem. Therefore, we design our dataset to cover different depth and width levels from 2 to 4, where each depth-width pair contains 100 questions, leading to a total of $900$ questions. We select the depth and width values to ensure that questions fit within the Qwen-2.5-32B-Instruct maximum token length and can be fully processed.

\textbf{Creative writing dataset.} Based on the common creative writing tasks \cite{yao2023tree}, we further propose a novel Depth-Width Writing (DW$^2$) benchmark, in which the task is to write a coherent essay using $K$ sentences with $K$ keywords drawn from across 23 major occupation groups~\cite{onet_taxonomy} in the United States. The \textbf{\textit{depth}} of the task is then captured by the number of sentences $K$, which dominates the length of the reasoning path. For \textbf{\textit{width}}, we use the \textit{normalized Shannon entropy} $\bar{H}(\mathcal{S})$ of the occupation categories in each keyword set $\mathcal{S} = \{c_1, c_2, ..., c_K\}$ with the following expression:
\begin{equation*}
\setlength{\abovedisplayskip}{4pt}
\setlength{\belowdisplayskip}{4pt}
    \bar{H}(\mathcal{S}) \;=\; \frac{H(\mathcal{S})}{H_{\max}} \;=\; -\,\frac{1}{H_{\max}}\,\sum_{i=1}^{K} p(c_i)\,\log_2 p(c_i),
\end{equation*}
where $H_{\max}=\log_2(K)$ is the maximum entropy when all keywords are from different categories, and $p(c_i) = \frac{1}{K}\sum_{j=1}^K \mathbbm{1}_{c_i = c_j} $ is the probability of category $c_i$ present in the set. Intuitively, a higher $\bar{H}(\mathcal{S})$ suggests a wider range of domain knowledge across occupation groups, hence making the completion of the task requiring broader knowledge. We then randomly sample 500 $K$-keyword sets for each $K\in\{4,8,12,16,20\}$, and then split them into 5 quintiles by their $\bar{H}$ scores, i.e., 100 questions per depth-width pair, which yields a total of $2,500$ questions. We select the depth and width values to ensure that any question fits within the Qwen-2.5-32B-Instruct maximum token length.

\begin{figure*}[t!]
\centering
    \begin{subfigure}[b]{\textwidth}\centering
        \includegraphics[width=\textwidth]{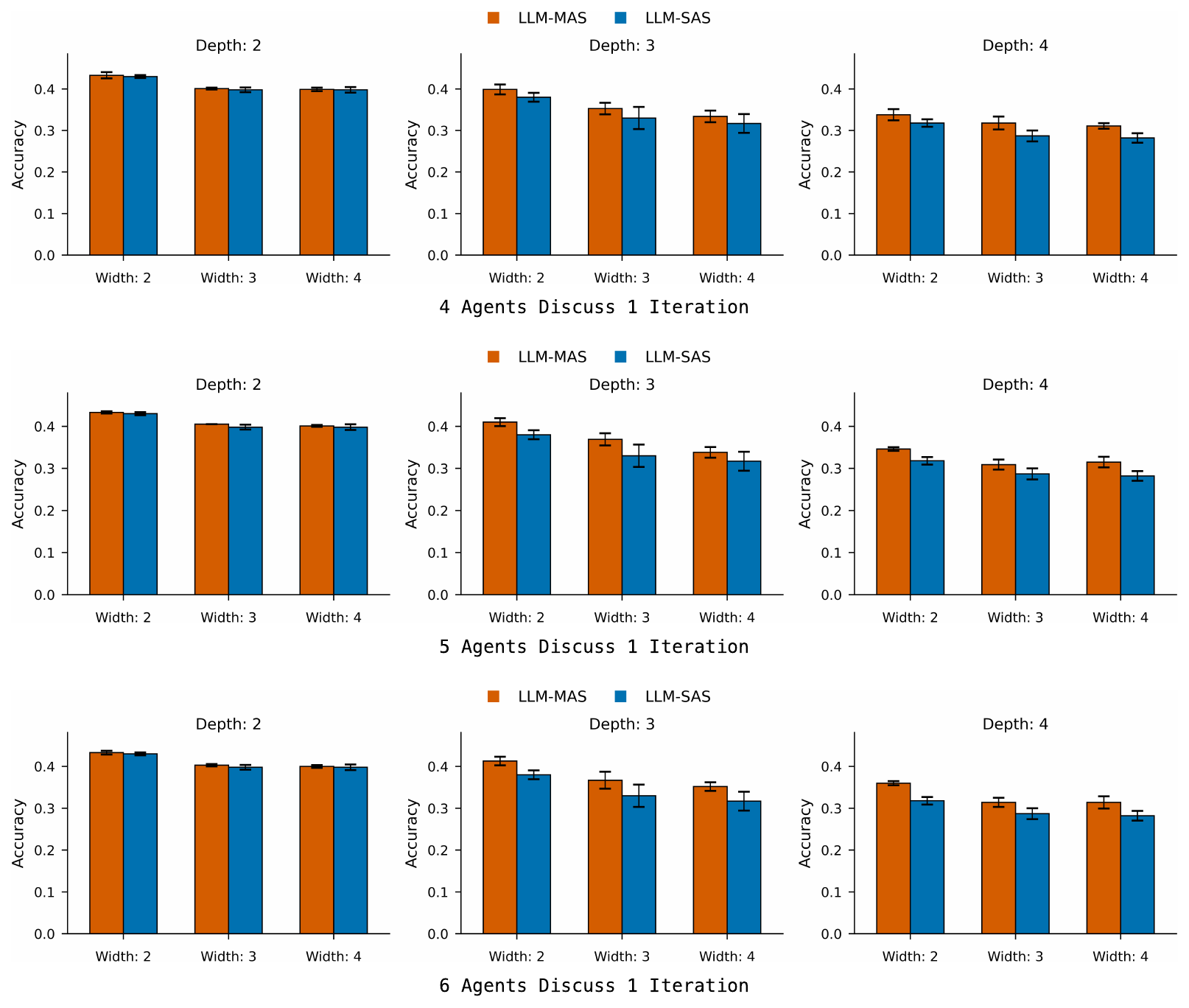}
        \caption{The accuracy of LLM-SAS and LLM-MAS.}
         \label{fig:math_acc}
    \end{subfigure}
    \hfill
    \begin{subfigure}[b]{\textwidth}\centering
        \includegraphics[width=\textwidth]{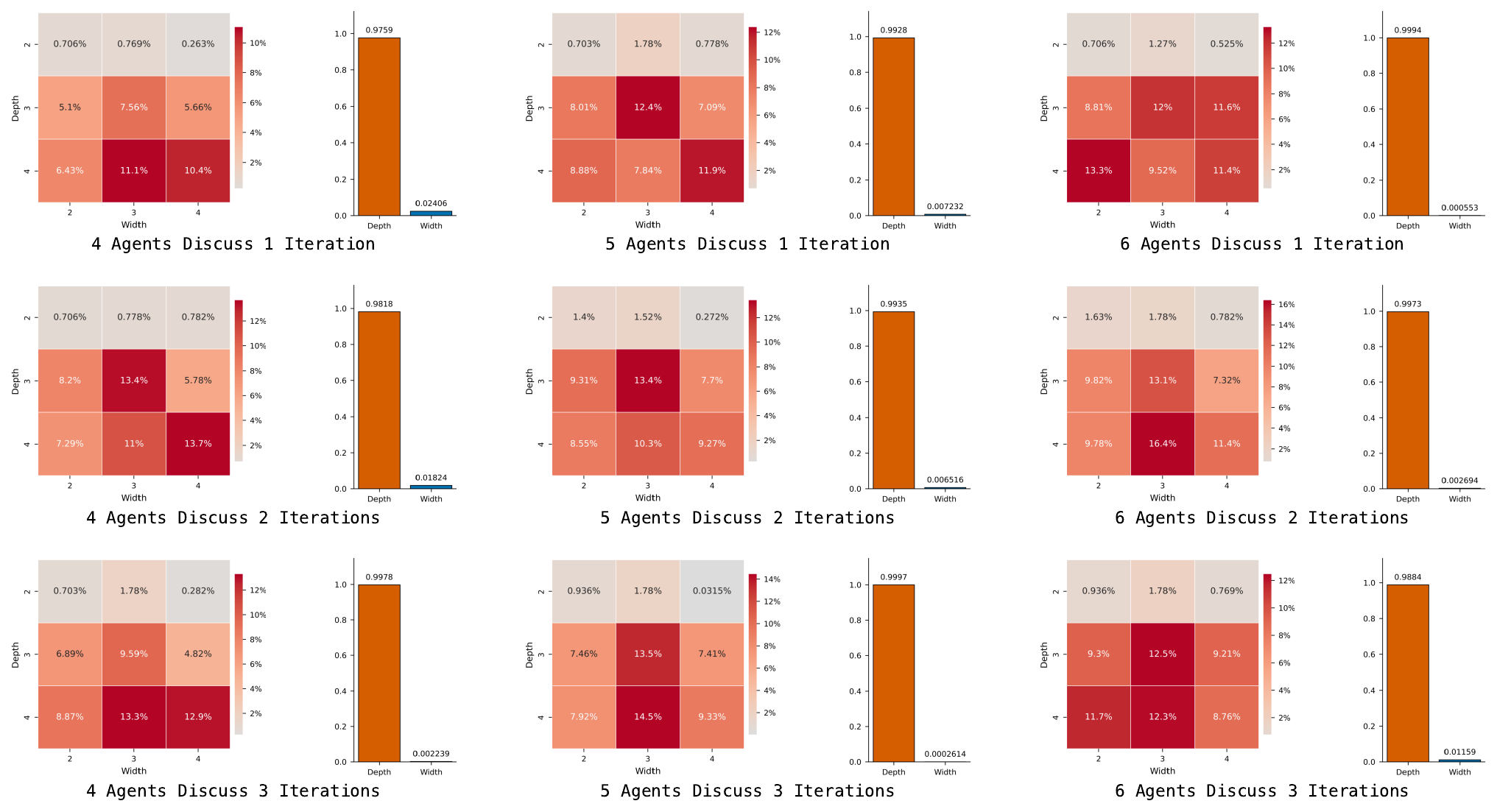}
        \caption{The performance gain and the S-Score.}
         \label{fig:math_gain}
    \end{subfigure}
\caption{Results on the math reasoning benchmark.}
\label{fig:math_results}
\end{figure*}

\textbf{Implementation.} For LLM-SAS, we implement the method based on the chain-of-thought (CoT) technique~\cite{wei2022chain}, where a single LLM is prompted to decompose each problem into a sequence of clear, step‑by‑step reasoning steps before reaching the final answer. For LLM-MAS, we implement the multi-agent debate framework~\cite{du2023improving}, which involves multiple LLM agents independently attempting a task and exchanging their reasoning along with their answers. After the discussion, a dedicated LLM agent will synthesize the information into a final response via summary prompting. For consistency, we use Qwen-2.5-32B-Instruct as the base model across all experiments. 

\textbf{Evaluation.} 
Since our goal is to compare the behaviours of LLM-SAS and LLM-MAS under different task complexities, we will focus on the performance gain (Definition \ref{def:pg}) at each depth-width pair and monitor how it changes along the depth and width dimensions. Specifically, we use the standard accuracy within each depth-width pair as the metric for the math reasoning task. For creative writing, we define a composite writing score, which is a product of a standard score ranging from 0 to 1 (e.g., completion within the target sentence count and inclusion of required keywords) and an LLM-generated quality score ranging from 0 to 10 (e.g., fluency, coherence, creativity). To quantify which dimension exerts greater influence on performance gain, we adopt a Shapley-$R^2$ decomposition~\cite{lipovetsky2001analysis}. Concretely, let $R^2(S)$ denote the coefficient of determination obtained from an ordinary least squares regression where the set of predictors is $S$. In our setting, the predictor set is $\{\text{depth}, \text{width}\}$. We therefore consider four regression fits: (i) the empty model $R^2(\emptyset)$, (ii) the depth-only model $R^2(\{\text{depth}\})$, (iii) the width-only model $R^2(\{\text{width}\})$, and (iv) the full model $R^2(\{\text{depth}, \text{width}\})$. The \textbf{Shapley score (S-Score)} for a predictor $x \in \{\text{depth}, \text{width}\}$ is defined as
\begin{equation*}
\setlength{\abovedisplayskip}{4pt}
\setlength{\belowdisplayskip}{4pt}
S(x) \;=\; \tfrac{1}{2}\Bigl(R^2(\{x\}) - R^2(\emptyset)\Bigr) \;+\; \tfrac{1}{2}\Bigl(R^2(\{\text{depth}, \text{width}\}) - R^2(\{\bar{x}\})\Bigr),
\end{equation*}
where $\bar{x}$ is the complementary predictor, e.g., if $x=\text{depth}$, then $\bar{x}=\text{width}$. The larger the S-Score for a dimension, the more dominant that dimension is considered in driving the performance gain.

%we employ a Shapley-\(R^2\) decomposition \cite{lipovetsky2001analysis}. By fitting regression models using depth-only, width-only, and both predictors, this method allocates the total explained variance (\(R^2\)) fairly between the two dimensions via Shapley score (S-Score) attribution: each dimension has an S-Score representing its average marginal contribution across all predictor inclusion orders. 

\subsection{Results and Analysis}

\textbf{Math reasoning.}  The results are summarised in Figure \ref{fig:math_results}. More results about the writing score are in the Appendix. We conduct experiments on an LLM-MAS with a varying number of agents from 4 to 6 (including a summarising agent)%, using a single discussion iteration
. Our findings reveal three key patterns that align with our theoretical analysis. First, as illustrated in Figure~\ref{fig:math_acc}, we observe that increasing task depth and width heightens task complexity, leading to a decrease in absolute performance for both single-agent and multi-agent systems. This validates our complexity measures as effective indicators of problem difficulty. Second, the heatmaps in Figure~\ref{fig:math_gain} show that the performance advantage of an LLM-MAS over a single-agent baseline (LLM-SAS) grows with task complexity; the collaborative benefit becomes more significant as problems become deeper and wider. Third, and most crucially, the bar charts within Figure~\ref{fig:math_gain} provide quantitative evidence that this advantage is more sensitive to increases in task depth than to comparable increases in task width. This result empirically confirms our theoretical analysis, highlighting that the depth of the task is a dominant factor in the efficacy of LLM-MAS.

\begin{figure*}[t!]
\centering
    \begin{subfigure}[b]{\textwidth}\centering
        \includegraphics[width=\textwidth]{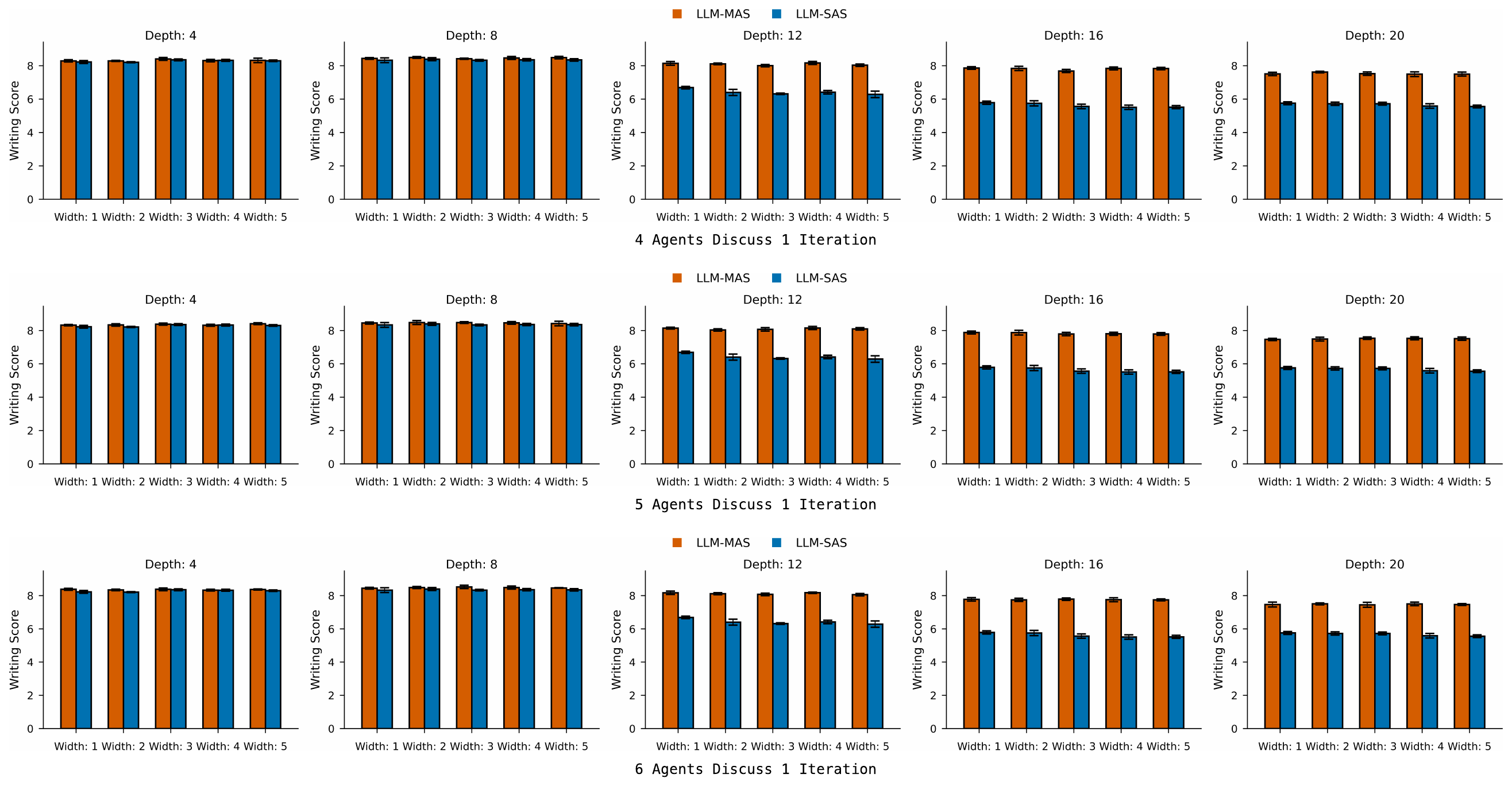}
        \caption{The writing score of LLM-SAS and LLM-MAS.}
         \label{fig:writing_score}
    \end{subfigure}
    \hfill
    \begin{subfigure}[b]{\textwidth}\centering
        \includegraphics[width=\textwidth]{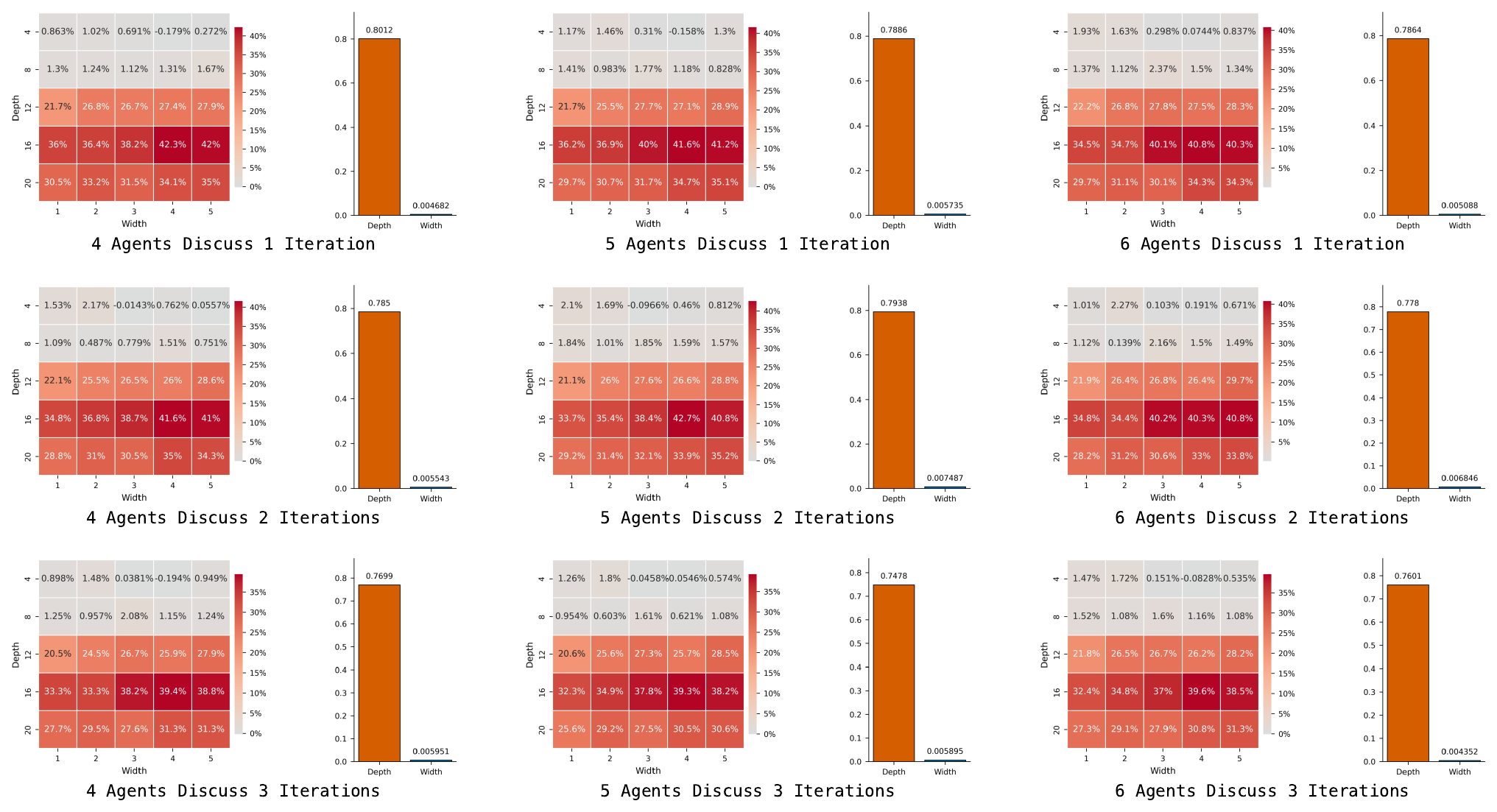}
        \caption{The performance gain and the S-Score.}
         \label{fig:writing_gain}
    \end{subfigure}
\caption{Results on the creative writing benchmark.}
\label{fig:writing_results}
\end{figure*}

\textbf{Creative writing.} The results are summarised in Figure \ref{fig:writing_results}. More results about the writing score are in the Appendix. We conduct experiments on the DW$^2$ benchmark, employing the same LLM-MAS setup with 4 to 6 agents% and a single discussion iteration
. The outcomes are highly consistent with those from the math reasoning experiments: the benefit of LLM-MAS over LLM-SAS grows with task complexity, and this gain is more pronounced with increases in depth than width. Notably, the magnitude of the performance improvement is substantially larger in this domain compared to the math reasoning task. We hypothesise that this is because creative writing is a generative task with a vast, open-ended solution space while satisfying a complex web of interacting constraints. As task complexity increases, we observe that the quality scores of LLM-MAS and LLM-SAS remain closely aligned, typically within a margin of about $5\%$, but LLM-SAS more frequently fail due to difficulties in meeting constraints, such as ensuring complete keyword coverage. LLM-MAS, however, excels by distributing this cognitive load across multiple agents and the debate process between them. Individual agents can propose solutions that champion different constraints, and the collaborative process allows for a more effective synthesis, leading to a final output that is both creative and holistically compliant.

% \begin{tcolorbox}[title=Clear performance gain from LLM-MAS across different \textit{\textbf{Depth}} levels]
% Some explanation of our observations on depth.
% \end{tcolorbox}

% \begin{tcolorbox}[title=Insignificant performance gain from LLM-MAS across different \textit{\textbf{Width}} levels]
% Some explanation of our observations on width.
% \end{tcolorbox}

% detailed analysis + this align with the math and our theory

% Experiments: 1. Get tasks with different difficulty-levels from dyval. 2. show the comparison between single agent and mac under tasks with different difficulty-levels.

% The interdependent problems needed to be solved for completing a single task (dyval depth and sequential tasks like smartphone operation) 

% Experiments: 1. Get tasks with different complexity-levels. 2. show the comparison between single agent and mac under tasks with different complexity-levels.

% the number of independent problems needed to be solved for completing a single task; creative writing with different domain knowledge; dyval changes the width of the task
\section{Challenges and Opportunities}

%\textbf{Evaluation of LLM-MAS necessity.} Existing evaluation metrics for LLM-MAS are predominantly grounded in empirical performance in downstream tasks, such as answer accuracy in mathematical reasoning, success rate in safety-critical reasoning~\cite{yang2025revisiting}, or heuristic-based evaluation in generative modeling~\cite{yu2025multi}. While this work attempts to characterise the complexity of the task itself, like empirical performance it still remains agnostic to the intrinsic properties of MAS themselves, such as model scale, agent capacity, and system complexity. Given that MAS typically incurs higher computational costs than single-agent systems, we argue that more comprehensive and system-aware evaluation metrics are needed to properly assess the necessity and efficiency of LLM-MAS.
\textbf{Evaluation of LLM-MAS necessity.} Current evaluation of LLM-MAS is dominated by empirical performance in downstream tasks, such as answer accuracy in mathematical reasoning, success rates in safety-critical reasoning~\cite{yang2025revisiting}, or heuristic-based assessments in generative modeling~\cite{yu2025multi}. While our work attempts to characterise task complexity as a complementary lens, this formulation, like pure performance metrics, remains agnostic to the intrinsic properties of MAS themselves, including model scale, agent capacity, and coordination overhead. Since LLM-MAS incur higher computational costs than LLM-SAS, a key challenge is to develop evaluation frameworks that jointly consider task complexity and system complexity, thereby providing a more principled measure of when the deployment of LLM-MAS is truly warranted.

\textbf{Dynamic task-oriented agent systems.} The configuration of MAS, including communication protocols, agent profiles, and agent capacities, is typically predetermined by system designers. This static and manually specified design often leads to limited adaptability, suboptimal performance, and significant inefficiencies in inference costs~\cite{yemas}. Our analysis of task complexity highlights why this is problematic: the effectiveness of MAS depends critically on the alignment between system configuration and the intrinsic demands of the task. A better understanding of how task complexity shapes the utility of higher-order interactions suggests that MAS should not be fixed, but rather adaptive. Building on this insight, we advocate for dynamic optimisation frameworks that allocate communication structures and agent capacities in response to the depth and width of the task at hand~\cite{ye2025xmasbuildingmultiagentsystems}. By tailoring MAS configurations to task complexity, such systems can improve both efficiency and task effectiveness, offering a principled pathway towards more adaptive and scalable LLM-based multi-agent architectures. %To address these challenges, we advocate for the development of adaptive optimisation frameworks that dynamically allocate MAS resources in response to task requirements~\cite{ye2025xmasbuildingmultiagentsystems}. Such systems should be capable of adjusting communication structures and agent capacities on demand, thereby improving both efficiency and task effectiveness.

\textbf{Datasets and benchmarks.} Analogous to human collaboration, MAS collaboration is most beneficial when tasks are inherently difficult, when individual expertise is insufficient, or when diverse perspectives are essential~\cite{yang2025revisiting}. However, most existing MAS benchmarks are adapted from single-agent settings and therefore fail to adequately capture these conditions. As a result, they often do not necessitate or reward genuine multi-agent collaboration, with only limited exceptions~\cite{ye2025xmasbuildingmultiagentsystems} that are explicitly designed for MAS. Looking forward, we argue that future benchmarks should be tailored to debate-style interactions, explicitly encoding the need for reasoning chains that benefit from critique, counter-arguments, and consensus-building. Such datasets would better reflect the unique challenges of LLM debate frameworks and enable systematic evaluation of when and how collaboration among agents translates into tangible performance gains. %To systematically evaluate LLM capabilities in multi-agent contexts, we emphasise the need for new benchmarks and datasets that explicitly encode collaboration requirements and reflect the unique challenges of MAS.

\textbf{Beyond depth and width.} Finally, while our analysis formalises task complexity in terms of depth (sequential reasoning steps) and width (capabilities required within a step), these are not the only possible dimensions. Alternative characterisations may consider, for example, the interaction complexity, which means the extent to which sub-tasks interdepend or conflict. Exploring these alternatives remains an open direction that could yield a richer taxonomy of complexity, bridging theoretical formulations with practical challenges faced by LLM-MAS. 

\section{Conclusion}

In this work, we present a theoretical and empirical investigation into the effectiveness of LLM multi-agent systems (LLM-MAS) over single-agent systems (LLM-SAS). To formulate our analysis, we introduce a novel measure of task complexity, defined in terms of reasoning sequentiality (depth) and knowledge diversity (width). We begin with a theoretical analysis, showing that while both dimensions yield a monotonically increasing performance gap between LLM-MAS and LLM-SAS, the gains from width are fundamentally bounded, whereas depth can provide unbounded improvements. We then validate these insights through empirical studies, demonstrating that: (1) both depth and width consistently enhance the relative performance of LLM-MAS over LLM-SAS, and (2) the benefits of increasing depth scale substantially, while the advantages from width quickly plateau once a finite bound is reached. Together, these results establish a principled foundation for the design of more efficient, effective, and theoretically grounded LLM-MAS. Our analysis focused on debate-based frameworks, which underpin many existing LLM-MAS. For instance, sub-task decomposition methods can be viewed as extensions of debate, tailored for project-level tasks where a large problem is divided into smaller questions, each solved by single agents or debate modules before being integrated into a coherent solution. Exploring such variants across different scenarios is non-trivial, as it requires not only generalising our theoretical results, but also designing new benchmarks with controllable depth and width for their specific settings. Similarly, extending our study to base models larger than Qwen-2.5-32B-Instruct would also be valuable but demands substantial computational resources. We therefore leave these as future work.

{
  \small
  \bibliographystyle{unsrt}
  \bibliography{neurips_2025}\label{reference}
}

\newpage
\appendix
\section{Proof for Proposition \ref{prop:grad_d_positive}}
\begin{proof}
By definition, we have:
\[
\Delta(d,w,N,r)\;\triangleq\;\frac{S_{\mathrm{multi}}-S_{\mathrm{single}}}{S_{\mathrm{single}}}
\;=\;\left(\frac{r\,[1-(1-s(w))^N]}{s(w)}\right)^{d}-1.
\]
Let \(s = s(w) = q^w \in (0,1)\) and define:
\[
A(s) \triangleq r\bigl[1 - (1 - s)^N\bigr], 
\quad f(s) \triangleq \frac{A(s)}{s}.
\]
Hence \(\Delta(d,w,N,r) = f(s)^d - 1\). Moreover, by assumption, we have \(f(s) = \frac{A(s)}{s} > 1\). Therefore, for the partial derivative in \(d\), we have 
\[
\frac{\partial\Delta}{\partial d} = \ln(f(s))\,f(s)^d > 0.
\]
For the partial derivative in \(w\), we have:
\[
\frac{\partial\Delta}{\partial w} = d\,f(s)^{d-1}\,f'(s)\,s'(w).
\]
Let \(t = 1 - s \in (0,1)\). Then, we have:
\[
A'(s)s - A(s)
= r\left[ Ns(1-s)^{N-1} - (1 - (1-s)^N) \right]
= -r\left(1 - t^N - N(1-t)t^{N-1}\right).
\]
By the Mean Value Theorem \cite{russell2020principles} applied to \(\varphi(x)=x^N\) on \([t,1]\), there exists \(\xi\in(t,1)\) such that
\[
1 - t^N = N \xi^{N-1}(1 - t) \ge N t^{N-1}(1 - t).
\]
Thus the bracketed expression is nonnegative, making \(A'(s)s - A(s) \le 0\), so \(f'(s)=\frac{A'(s)s - A(s)}{s^2}\le 0\). Since \(s'(w) = q^w \ln q < 0\), it follows that
\[
\frac{\partial\Delta}{\partial w} = d\,f(s)^{d-1}\,f'(s)\,s'(w) > 0.
\]

\end{proof}

\section{Proof for Proposition \ref{prop:limits_depth_width}}
\begin{proof}
By definition, we have:
\[
\Delta(d,w,N,r)\;\triangleq\;\frac{S_{\mathrm{multi}}-S_{\mathrm{single}}}{S_{\mathrm{single}}}
\;=\;\left(\frac{r\,[1-(1-s(w))^N]}{s(w)}\right)^{d}-1.
\]
Let \(s = s(w) = q^w \in (0,1)\) and define:
\[
A(s) \triangleq r\bigl[1 - (1 - s)^N\bigr], 
\quad f(s) \triangleq \frac{A(s)}{s}.
\]
As \(w \to \infty\), \(s(w) = q^w \to 0\). By the Binomial Theorem \cite{russell2020principles}, for integer \(N\), we have $(1 - s)^N = 1 - N s + \binom{N}{2} s^2 - \dots + (-s)^N$. Therefore, we can have the following representation: $1 - (1 - s)^N = N s + O(s^2).$ Therefore, we can have:
\[
\lim_{w\to\infty} \Delta(d,w,N,r) = (rN)^d - 1.
\]
Since \(f(s)>1\), \(f(s)^d \to +\infty\) as \(d \to +\infty\), therefore we can have:
\[
\lim_{d\to\infty}\Delta(d,w,N,r) = +\infty.
\]
\end{proof}

\section{Additional Experimental Results}

\begin{figure*}[t]
    \centering
    \includegraphics[width=.9\linewidth]{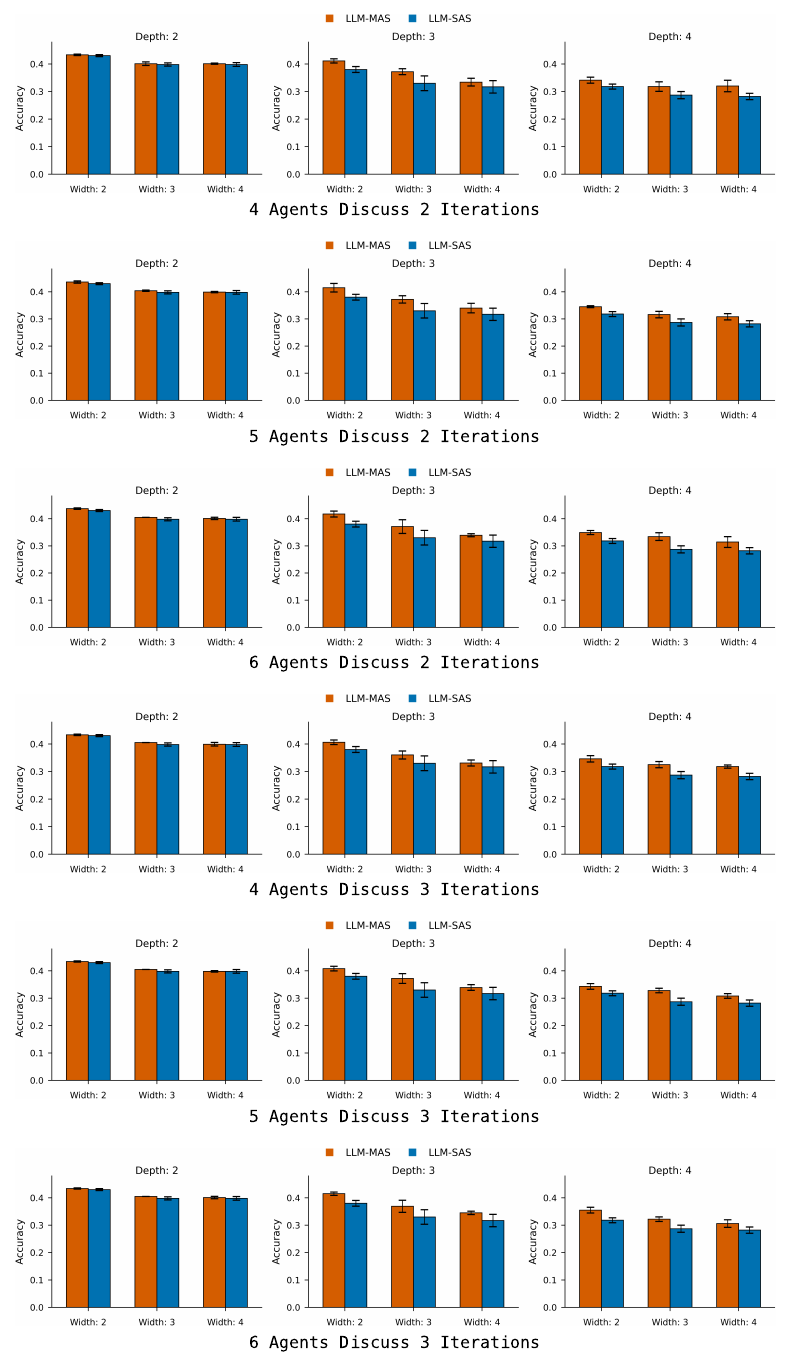}
    \caption{The accuracy of LLM-SAS and LLM-MAS on math reasoning.}
\end{figure*}

\begin{figure*}[t]
    \centering
    \includegraphics[width=\linewidth]{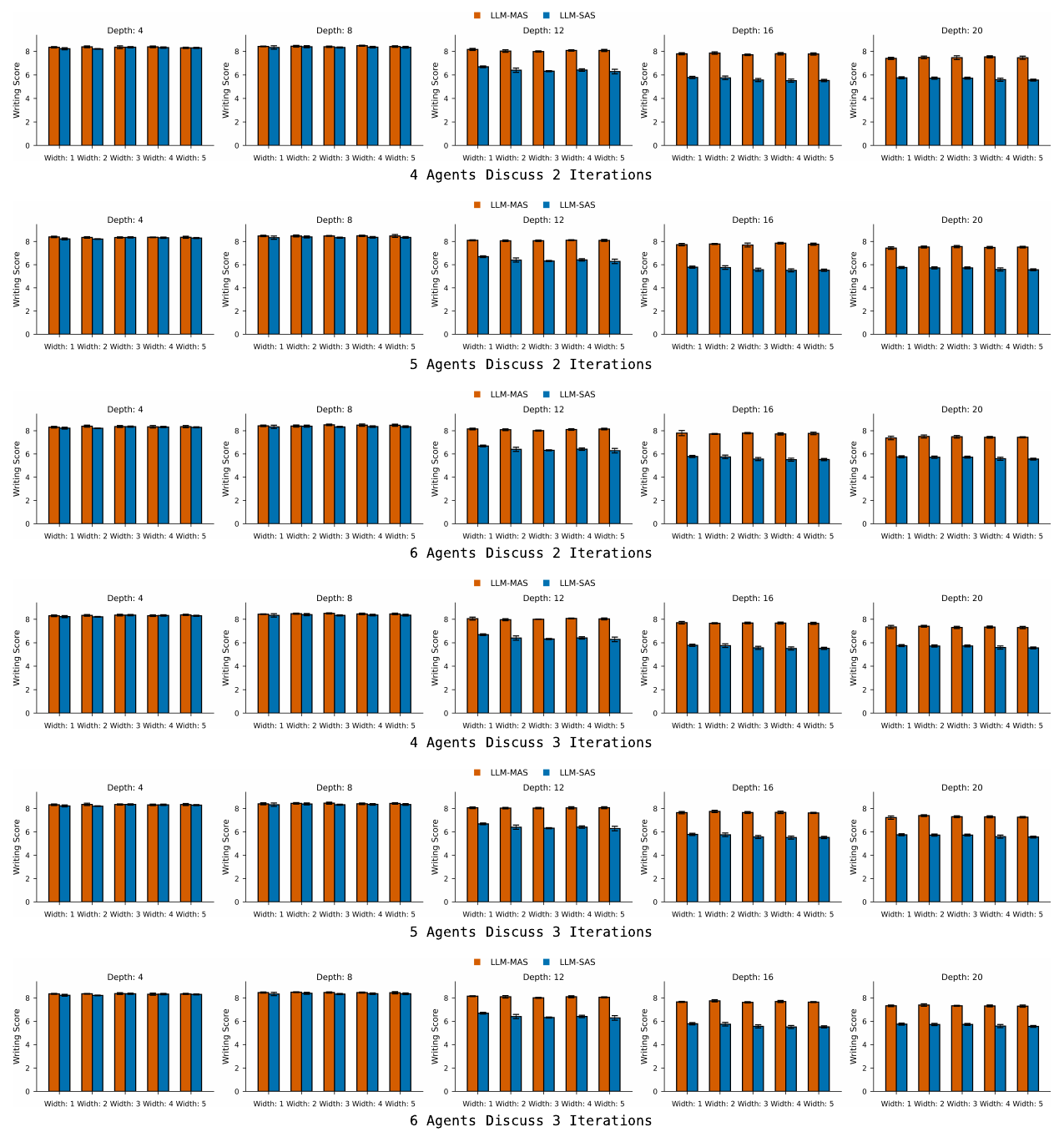}
    \caption{The writing score of LLM-SAS and LLM-MAS on creative writing.}
\end{figure*}

\end{document}